%% file: main.tex
\documentclass[10pt,twocolumn,letterpaper]{article}

\usepackage[pagenumbers]{cvpr} 

\input{preamble}

\definecolor{cvprblue}{rgb}{0.21,0.49,0.74}
\usepackage[pagebackref,breaklinks,colorlinks,allcolors=cvprblue]{hyperref}

\def\paperID{8} 
\def\confName{CVPR}
\def\confYear{2026}

\title{Masked Language Prompting for Generative Data Augmentation in Few-shot Fashion Style Recognition}

\author{
Yuki Hirakawa\\ ZOZO Research\\ {\tt\small yuki.hirakawa@zozo.com} \and Ryotaro Shimizu\\ ZOZO Research\\ {\tt\small ryotaro.shimizu@zozo.com}
}

\begin{document}
\maketitle
\input{sec/0_abstract}    
\input{sec/1_intro}
\input{sec/2_methodology}
\input{sec/3_experiments}
\input{sec/4_conclusion}
{
    \small
    \bibliographystyle{ieeenat_fullname}
    \bibliography{main}
}


\end{document}

%% file: preamble.tex
\usepackage{bm}
\usepackage{multirow}
\usepackage{array}
\usepackage{colortbl}
\usepackage{tcolorbox}
\usepackage{xcolor}



%% file: sec/0_abstract.tex
\begin{abstract}
Constructing datasets for fashion style recognition is challenging due to the subjectivity and ambiguity of style concepts. Recent text-to-image (T2I) models enable generative data augmentation, but existing methods conditioned on class names or captions often fail to balance visual diversity and fidelity. We propose Masked Language Prompting (MLP), which masks words in a reference caption and uses large language models to generate diverse yet coherent completions. These prompts guide T2I models to synthesize labeled images without fine-tuning, making MLP suitable for trend-sensitive concepts such as fashion styles. Experiments on FashionStyle14 show that MLP outperforms class-name and caption-based baselines under limited supervision. Quantitative and qualitative analyses demonstrate that MLP produces substantially more diverse images, validating its effectiveness as a data augmentation strategy.
\end{abstract}

%% file: sec/1_intro.tex
\section{Introduction}
\label{sec:intro}
Fashion style recognition~\cite{Kiapour2014HipsterWD, TakagiICCVW2017, simo2016fashion, an2023conceptual} is an important computer vision task with applications in recommendation and trend analysis. However, it remains challenging due to the complex interplay of visual attributes such as color, shape, and fabric, as well as the subjectivity and cultural dependence of style interpretation~\cite{an2023conceptual}. These factors make large-scale annotation costly and difficult~\cite{TakagiICCVW2017, an2023conceptual}, leaving practitioners to train models from limited labeled data.
\begin{figure}[t]
    \centering
    \begin{minipage}[b]{0.3\linewidth}
        \centering
        \includegraphics[width=\linewidth]{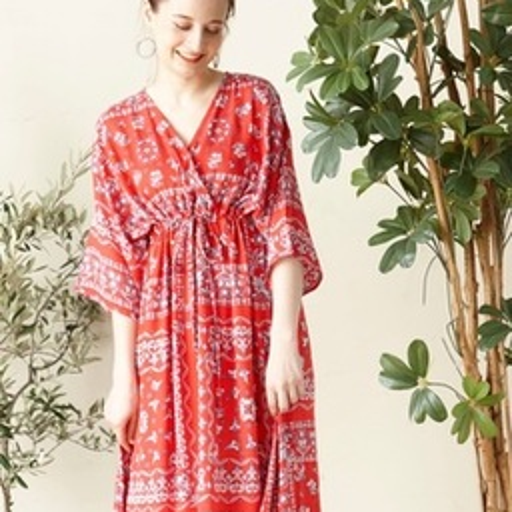}
        \subcaption{Original}
        \label{fig:aug_original}
    \end{minipage}
    \hspace{0.2cm}
    \begin{minipage}[b]{0.3\linewidth}
        \centering
        \includegraphics[width=\linewidth]{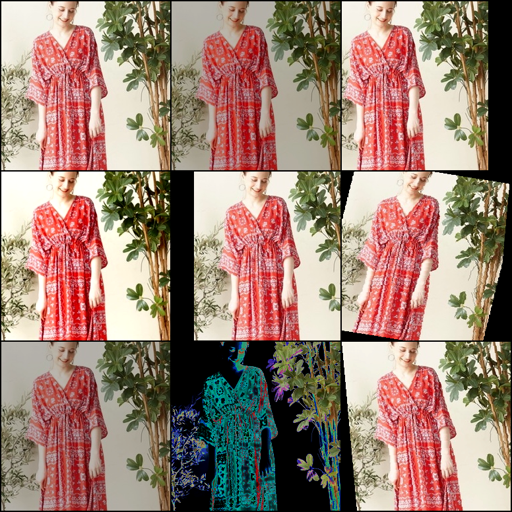}
        \subcaption{RandAug.}
        \label{fig:aug_rand}
    \end{minipage}
    \hspace{0.2cm}
    \begin{minipage}[b]{0.3\linewidth}
        \centering
        \includegraphics[width=\linewidth]{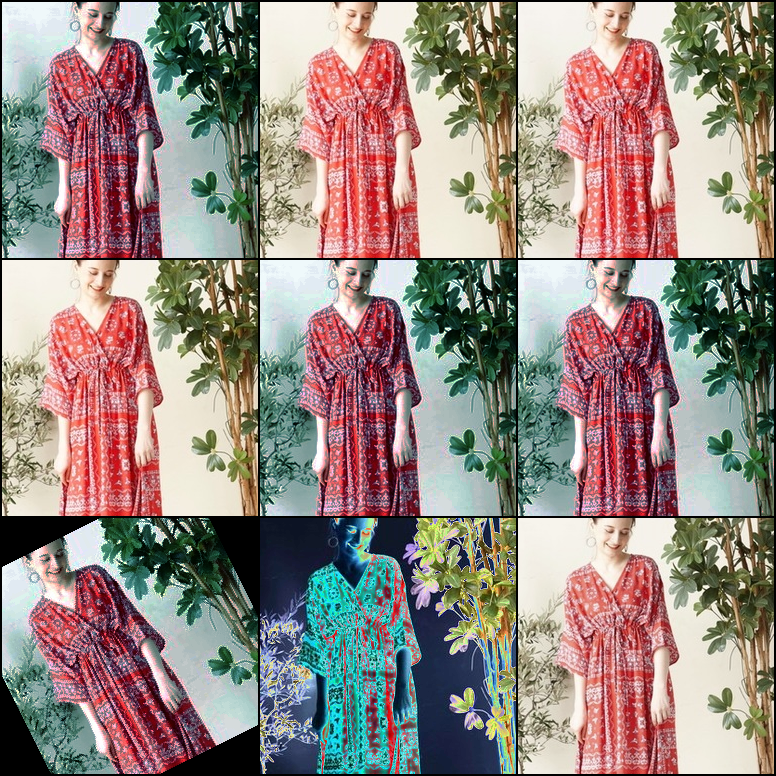}
        \subcaption{AutoAug.}
        \label{fig:aug_auto}
    \end{minipage}
    \begin{minipage}[b]{0.3\linewidth}
        \centering
        \includegraphics[width=\linewidth]{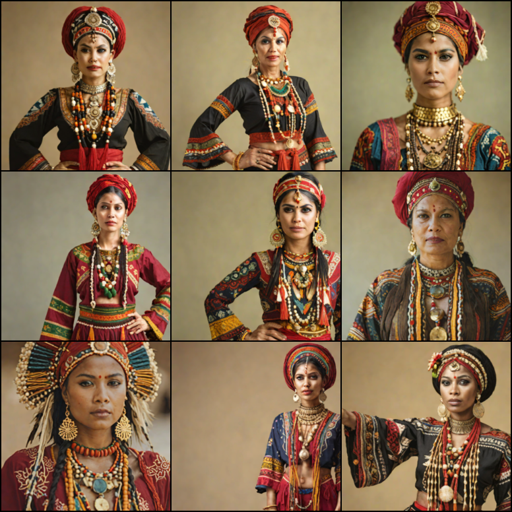}
        \subcaption{Class}
        \label{fig:aug_cls}
    \end{minipage}
    \hspace{0.2cm}
    \begin{minipage}[b]{0.3\linewidth}
        \centering
        \includegraphics[width=\linewidth]{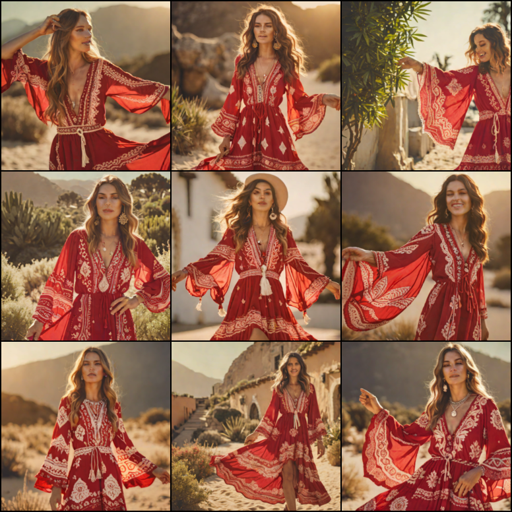}
        \subcaption{Caption}
        \label{fig:aug_caption}
    \end{minipage}
    \hspace{0.2cm}
    \begin{minipage}[b]{0.3\linewidth}
        \centering
        \includegraphics[width=\linewidth]{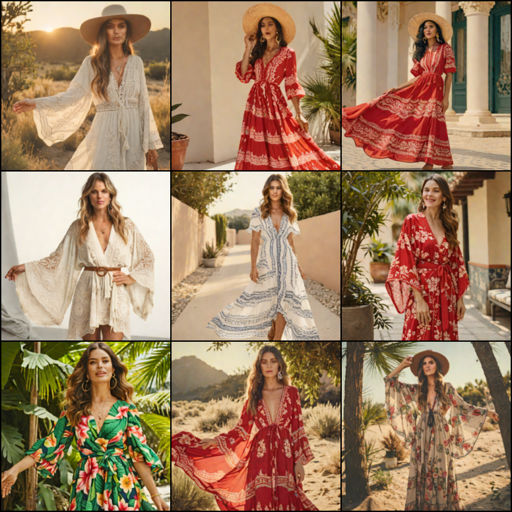}
        \subcaption{MLP (ours)}
        \label{fig:aug_mlp}
    \end{minipage}
    \caption{\textbf{Comparison of data augmentation methods.} (a) shows the reference real image. (b) and (c) correspond to conventional pixel-level augmentations. (d), (e), and (f) illustrate generative augmentations based on the class label, the caption of the reference image, and our proposed MLP.}
    \label{fig:augmentation}
\end{figure}
\begin{figure*}[t]
    \centering
    \includegraphics[width=1.0\linewidth]{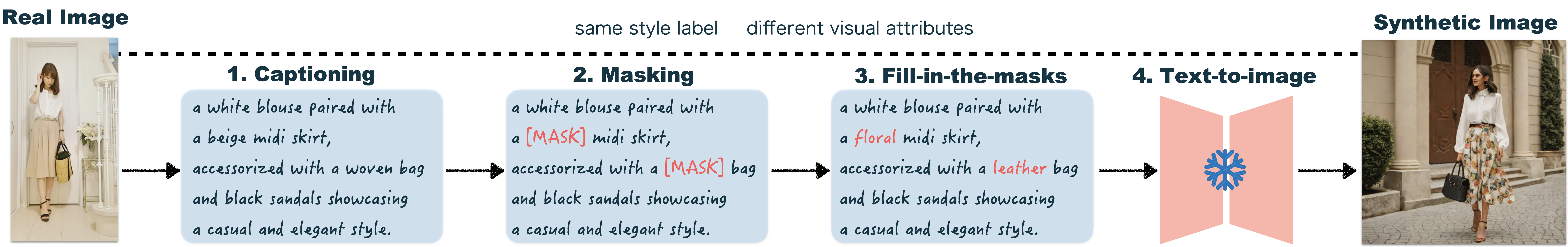}
    \caption{\textbf{Masked language prompting.}  Our augmentation pipeline consists of four steps: (1) \emph{captioning}, (2) \emph{masking}, (3) \emph{fill-in-the-masks}, and (4) \emph{text-to-image generation}. This process enables effective synthesis of images that preserve the style semantics of the original reference while introducing attribute-level variations without fine-tuning T2I models.}
    \label{fig:mlp}
\end{figure*}

To mitigate data scarcity, conventional image-based augmentation methods—such as geometric and photometric transformations—are widely used. However, these pixel-level operations provide limited semantic variation (Fig.~\ref{fig:aug_rand}, Fig.~\ref{fig:aug_auto}), restricting their ability to model style-related differences. Recent advances in text-to-image (T2I) models~\cite{rombach2022high, Ramesh2022HierarchicalTI} enable generative augmentation~\cite{kim2024datadream, he2023is, zhou2023synthetic, dunlap2023diversify, zhang2023expanding, wang2024attributed}, which synthesizes semantically rich training samples from textual prompts. While promising for object and scene recognition, its applicability to subjective concepts such as fashion styles remains underexplored. In this work, we investigate T2I-based generative augmentation to improve few-shot fashion style recognition by generating diverse yet style-consistent images from limited labeled data.

Existing generative augmentation approaches typically rely on class-name prompts (e.g., “A photo of a woman wearing a \texttt{[CLASS]} style outfit”), which often produce semantically misaligned results for culturally grounded and polysemous styles. For example, Japanese ethnic style is characterized by traditional patterns and loose silhouettes (Fig.~\ref{fig:aug_original}), whereas class-name prompts generate culturally divergent outputs such as saris or Middle Eastern garments (Fig.~\ref{fig:aug_cls}). Using reference captions as prompts (e.g., “A photo of a woman wearing \texttt{[CAPTION]}”) improves fidelity but yields limited diversity (Fig.~\ref{fig:aug_caption}).

To balance fidelity and diversity, we propose Masked Language Prompting (MLP), which selectively masks words in reference captions and leverages large language models (LLMs) to generate diverse yet semantically coherent prompt variants. These prompts guide T2I models to synthesize labeled images without fine-tuning. We evaluate MLP on the FashionStyle14~\cite{TakagiICCVW2017} dataset by training linear classifiers on CLIP~\cite{radford2021learning} embeddings using a small set of real images augmented with synthetic data. Experimental results show that MLP consistently outperforms class-name and caption-based baselines, demonstrating its effectiveness for data augmentation in conceptually ambiguous tasks such as fashion style recognition.

%% file: sec/2_methodology.tex
\section{Methodology}
\label{sec:methodology}
\subsection{Generative Data Augmentation}
Suppose we have a small labeled dataset $\mathcal{D}_{r}=\{(\bm{x}_i, y_i)\}_{i=1}^{n_r}$, where $\bm{x}_i\in\mathbb{R}^{d}$ denotes a real image and $y_i\in\{1,\ldots,C\}$ is its class label for $C$ categories. To mitigate overfitting due to data scarcity, we introduce a set of synthetic labeled images $\mathcal{D}_{s}=\{(\bm{x}^{\prime}_i, y^{\prime}_i)\}_{i=1}^{n_s}$ generated via generative models, where $\bm{x}^{\prime}_i\in\mathbb{R}^{d}$ denotes a synthetic image and $y^{\prime}_i\in\{1,\ldots,C\}$ is its class label. The extended dataset $\mathcal{D}_{e} = \mathcal{D}_r \cup \mathcal{D}_s$ is used to train a classifier $F_{\Theta}$ by minimizing the sum of cross-entropy losses $\mathcal{L}_{\textrm{ce}}$ over real and synthetic mini-batches: \begin{equation} \mathcal{L}(F_{\Theta}, \mathcal{B}_{r} \cup \mathcal{B}_{s}) = \mathcal{L}_{\textrm{ce}}(F_{\Theta}, \mathcal{B}_r) + \mathcal{L}_{\textrm{ce}}(F_{\Theta}, \mathcal{B}_s), \label{eq:loss} \end{equation} where $\mathcal{B}_r \subset \mathcal{D}_r$ and $\mathcal{B}_s \subset \mathcal{D}_s$ denote mini-batches from real and synthetic data, respectively.
\subsection{Masked Language Prompting}
Effective generative augmentation demands synthetic images that preserve stylistic fidelity while exhibiting sufficient diversity. To address this issue, we propose Masked Language Prompting (MLP), a strategy that generates diverse yet semantically consistent textual prompts by masking parts of reference captions and completing them with LLMs. As illustrated in Fig.~\ref{fig:mlp}, given a labeled sample $(\bm{x}, y) \in \mathcal{D}_r$, our pipeline consists of four steps: (1) generating a detailed caption using LLMs that describes visual attributes such as color, garment type, and style; (2) identifying and masking key attribute words (\eg color, material, category); (3) completing the masked caption with contextually appropriate alternatives via LLMs; and (4) synthesizing an image from the completed prompt using T2I models. The resulting synthetic image $\bm{x}^\prime$ retains the style semantics $y$ of the original image $\bm{x}$ while introducing controlled attribute-level variations, thereby enhancing data diversity without compromising label consistency. We design the prompt for the \emph{captioning} step to elicit fine-grained descriptions of the clothing worn by the person in the image, covering attributes such as color, category, and design~\cite{XuewenECCV20Fashion}. In the \emph{fill-in-the-masks} step, we design the prompt to accept a partially masked caption as a placeholder and use the LLMs to complete it in a contextually appropriate manner, resulting in a natural variant of the original description. A key advantage of MLP is that it leverages the prior knowledge of LLMs to generate task-relevant augmentations without requiring fine-tuning of T2I models.

%% file: sec/3_experiments.tex
\section{Experiments}
\label{sec:3_experiments}
\begin{table*}[t]
    \centering
    \begin{tabular}{c|ccccccc|c}
    \toprule
       Shot & Method & N/A & RandAug. & AutoAug. & CutMix & Mixup & AugMix & Avg.\\
    \midrule
    \multirow{4}{*}{$1$} 
    & N/A & $\underline{0.347}$        & 0.362        & 0.351        & \textcolor{blue}{0.346}        & \textcolor{blue}{0.320}        & 0.368 & 0.349\\
    & Class       & 0.366        & 0.393        & 0.376        & 0.394        & 0.370        & 0.389 & 0.381\\
    & Caption   & 0.405        & 0.425        & 0.423        & 0.415        & 0.390        & 0.422& 0.413 \\
    & MLP (ours) & $\bm{0.424}$ & $\bm{\textcolor{red}{0.447}}$ & $\bm{0.445}$ & $\bm{0.432}$ & $\bm{0.410}$ & $\bm{\textcolor{red}{0.447}}$ & $\bm{0.434}$\\
    \midrule
    \multirow{4}{*}{$2$} 
    & N/A & $\underline{0.471}$        & 0.510        & 0.485        & 0.493        & 0.472        & 0.490 & 0.487\\
    & Class       & 0.472        & 0.513        & 0.492        & 0.511        & 0.484        & 0.500 & 0.495\\
    & Caption   & 0.537        & 0.559        & 0.545        & 0.542        & 0.533        & 0.551 & 0.545\\
    & MLP (ours) & $\bm{0.551}$ & $\bm{\textcolor{red}{0.568}}$ & $\bm{0.559}$ & $\bm{0.555}$ & $\bm{0.546}$ & $\bm{0.560}$ & $\bm{0.557}$\\
    \midrule
    \multirow{4}{*}{$4$} 
    & N/A & $\underline{0.588}$        & 0.607        & 0.590        & 0.592        & \textcolor{blue}{0.577}        & 0.594 & 0.591\\
    & Class       & \textcolor{blue}{0.565}        & 0.596        & \textcolor{blue}{0.569}        & \textcolor{blue}{0.586}        & \textcolor{blue}{0.571}        & \textcolor{blue}{0.581} & 0.578\\
    & Caption   & 0.627        & 0.645        & 0.640        & 0.629        & 0.620        & 0.638 & 0.633\\
    & MLP (ours) & $\bm{0.630}$ & $\bm{\textcolor{red}{0.651}}$ & $\bm{0.647}$ & $\bm{0.640}$ & $\bm{0.628}$ & $\bm{0.640}$ & $\bm{0.639}$\\
    \midrule
    \multirow{4}{*}{$8$} 
    & N/A & $\underline{0.677}$        & 0.698        & 0.679        & 0.689        & \textcolor{blue}{0.672}        & 0.681 & 0.682\\
    & Class       & \textcolor{blue}{0.636}        & \textcolor{blue}{0.655}        & \textcolor{blue}{0.637}        & \textcolor{blue}{0.655}        & \textcolor{blue}{0.631}        & \textcolor{blue}{0.651} & 0.644\\
    & Caption   & 0.698        & 0.706        & 0.700        & 0.703        & 0.693        & 0.704 & 0.701\\
    & MLP (ours) & $\bm{0.703}$ & $\bm{\textcolor{red}{0.714}}$ & $\bm{0.706}$ & $\bm{0.712}$ & $\bm{0.696}$ & $\bm{0.709}$ & $\bm{0.707}$\\
    \midrule
    \multirow{4}{*}{$16$} 
    & N/A & $\underline{0.739}$        & $\bm{\textcolor{red}{0.760}}$ & 0.745        & 0.752        & 0.744        & 0.753 & 0.749\\
    & Class       & \textcolor{blue}{0.696}        & \textcolor{blue}{0.702}        & \textcolor{blue}{0.687}        & \textcolor{blue}{0.712}        & \textcolor{blue}{0.678}        & \textcolor{blue}{0.701} & 0.696 \\
    & Caption   & 0.751        & 0.753        & 0.752        & 0.752        & 0.743        & 0.750 & 0.750\\
    & MLP (ours) & $\bm{0.755}$ & 0.759        & $\bm{0.758}$ & $\bm{0.756}$ & $\bm{0.745}$ & $\bm{0.759}$ & $\bm{0.755}$\\
    \bottomrule
    \end{tabular}
    \caption{\textbf{Mean classification accuracy of linear classifiers trained with different data augmentation methods for each shot.} Rows correspond to generative augmentation settings, where ``N/A'' denotes the use of real images only, while columns indicate standard augmentation settings, with ``N/A'' denotes no standard augmentation applied. \underline{Underlined values} indicate (1) augmentation-free baselines trained on real data without augmentation. \textbf{Bold} highlights (2) the best result within each shot and prompting methods. \textcolor{red}{Red} marks (4) the best overall result for each shot. \textcolor{blue}{Blue} denotes (3) performance lower than the augmentation-free baseline. }
    \label{tab:main_result}
\end{table*}
\begin{figure*}[t]
\centering
\begin{tabular}{m{3cm}|m{3cm}m{3cm}|m{3cm}m{3cm}}
\toprule
\multicolumn{1}{c|}{Reference} & \multicolumn{2}{c|}{Synthetic images / Success} & \multicolumn{2}{c}{Synthetic images / Failure}\\
\midrule
\includegraphics[width=3cm]{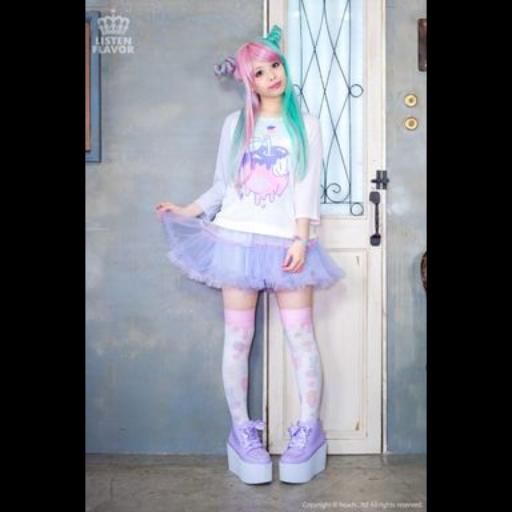} & 
\includegraphics[width=3cm]{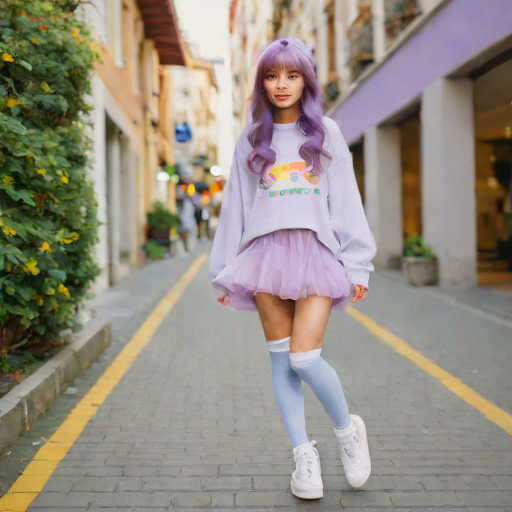} & 
\includegraphics[width=3cm]{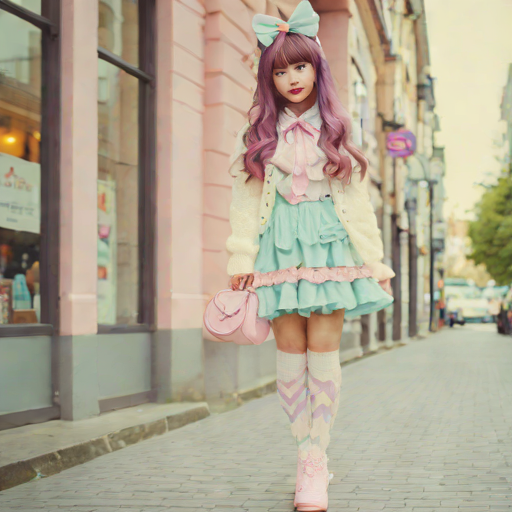} & 
\includegraphics[width=3cm]{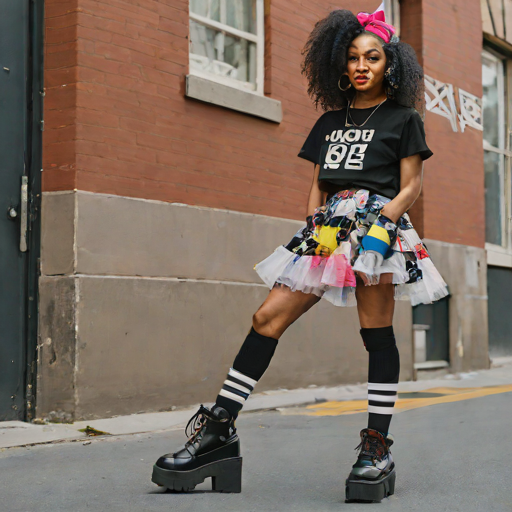} &
\includegraphics[width=3cm]{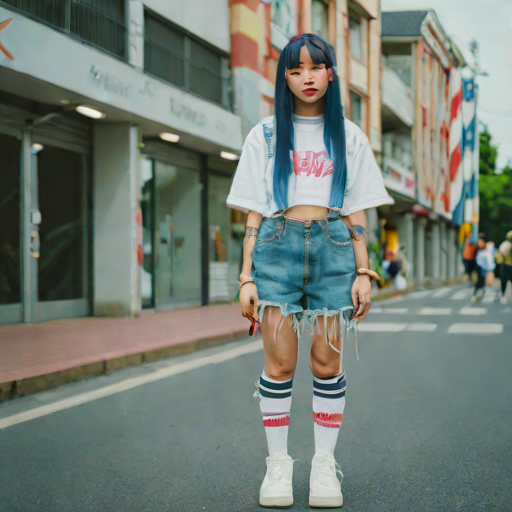}\\
\midrule
 \fontsize{6pt}{6pt}\selectfont{\emph{a pastel-themed outfit with a white graphic top, lavender tutu skirt, patterned knee-high socks, and platform shoes, embodying a kawaii Harajuku fashion style.}} & 
 \fontsize{6pt}{6pt}\selectfont{\emph{a pastel-themed \textcolor{red}{oversized sweater} with a \textcolor{red}{cute} graphic print, lavender tutu skirt, patterned knee-high socks, and platform shoes, embodying a kawaii \textcolor{red}{street} fashion style.}} & 
 \fontsize{6pt}{6pt}\selectfont{\emph{a pastel-themed \textcolor{red}{dress} with a \textcolor{red}{fluffy cardigan}, \textcolor{red}{oversized hair bow}, \textcolor{red}{cute handbag}, patterned knee-high socks, and platform shoes, embodying a kawaii Harajuku fashion style.}} & 
 \fontsize{6pt}{6pt}\selectfont{\emph{a \textcolor{red}{streetwear} outfit with a \textcolor{red}{bold} graphic \textcolor{red}{tee}, \textcolor{red}{layered} tutu skirt, patterned knee-high socks, and platform shoes, embodying a \textcolor{red}{playful urban aesthetic}.}} & 
\fontsize{6pt}{6pt}\selectfont{\emph{a \textcolor{red}{streetwear} outfit with a white graphic top, \textcolor{red}{oversized denim shorts}, patterned knee-high socks, and platform shoes, embodying a \textcolor{red}{playful} Harajuku aesthetic style.}}\\
\bottomrule
\end{tabular}
\caption{\textbf{Visualization of image-caption pairs generated using MLP, including both successful and failure cases.} The leftmost column shows a reference \emph{fairy} style image with a GPT-generated caption. Columns 2-5 display image-caption pairs generated by MLP conditioned on the reference, where columns 2-3 successfully capture key characteristics of the \emph{fairy} style—\eg, aesthetics of cuteness, delicacy, and soft color tones—while columns 4-5 illustrate failure cases that instead exhibit traits more aligned with the \emph{street} style, such as a rough appearance and a casual silhouette. Words highlighted in \textcolor{red}{red} indicate those completed during the \emph{fill-in-the-masks} step.}
\label{fig:mlp_visualization_1}
\end{figure*}
\begin{figure}[t]
    \centering
    \begin{minipage}[b]{0.45\linewidth}
        \centering
        \includegraphics[width=\linewidth]{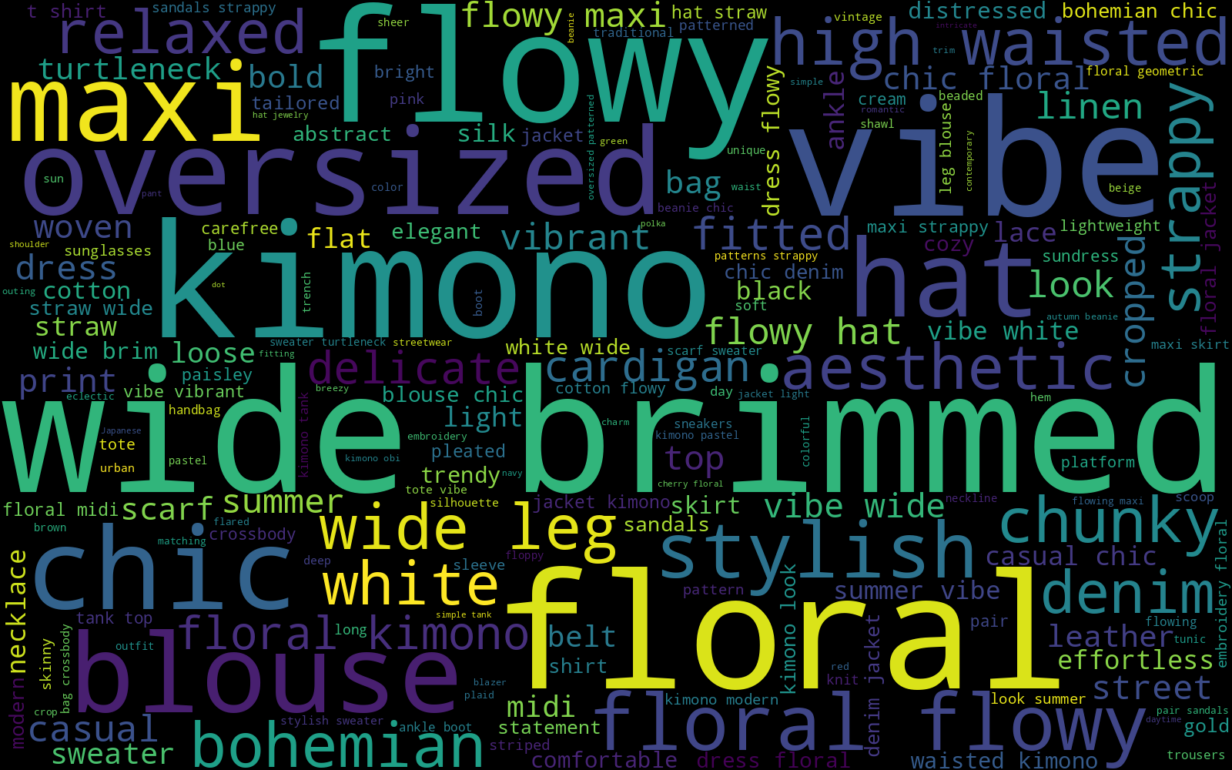}
        \subcaption{\emph{ethnic}}
    \end{minipage}
    \begin{minipage}[b]{0.45\linewidth}
        \centering
        \includegraphics[width=\linewidth]{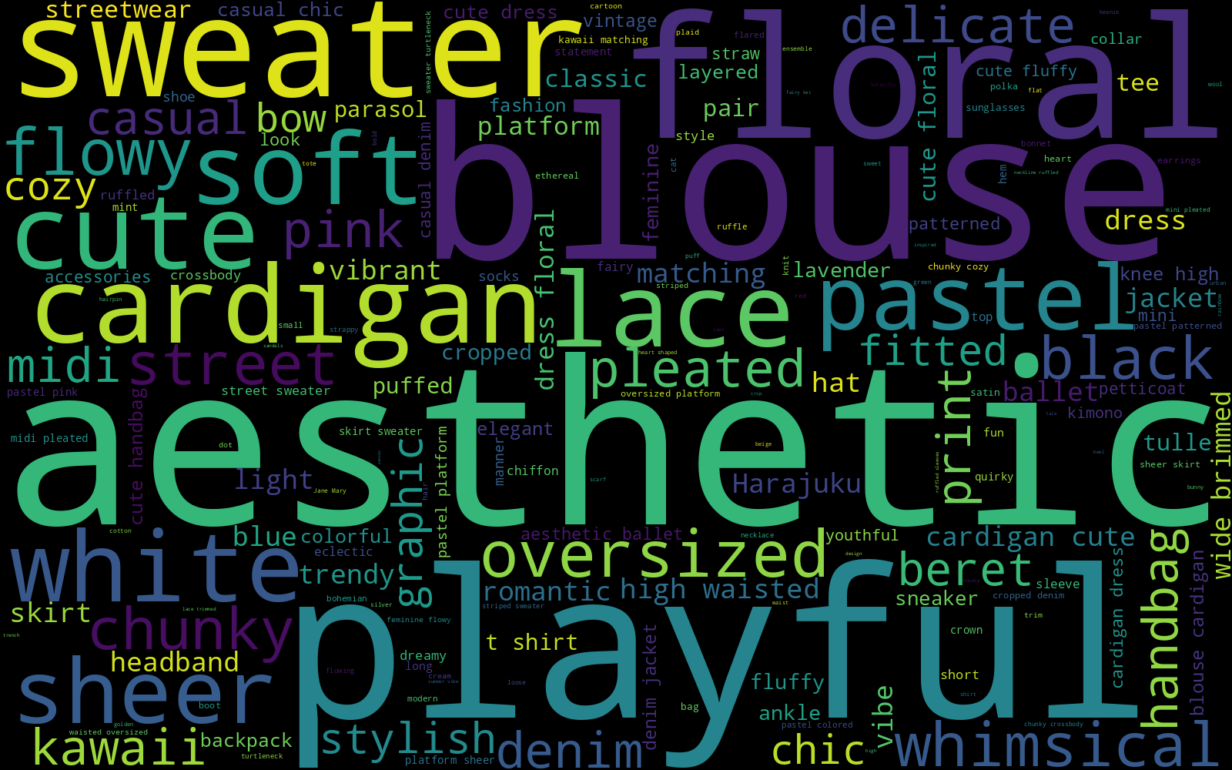}
        \subcaption{\emph{fairy}}
    \end{minipage}
    \caption{\textbf{Wordclouds of LLM-generated completions for masked captions.} Words are aggregated per class from LLMs outputs under the $n_{\mathrm{shot}} = 16$ setting, showing how \texttt{[MASK]} tokens are replaced during \emph{fill-in-the-masks} step.}
    \label{fig:wordcloud}
\end{figure}
\noindent\textbf{Data Augmentation.} We compare our MLP with standard augmentations, including RandAugment~\cite{cubuk2020randaugment}, AutoAugment~\cite{cubuk2019autoaugment}, CutMix~\cite{yun2019cutmix}, Mixup~\cite{zhang2018mixup}, and AugMix~\cite{hendrycks2020augmix}. For generative augmentation, we compare three prompt types—Class, Caption, and our MLP—following the formats: ``\emph{A photo of a woman wearing a} \texttt{[CLASS]} \emph{style outfit.},'' ``\emph{A photo of a woman wearing} \texttt{[CAPTION]}.'' and ``\emph{A photo of a woman wearing} $G_{\mathrm{MLP}}(\texttt{[CAPTION]})$.'', respectively. Here, $G_{\mathrm{MLP}}$ masks part of a caption and completes it with LLMs. Both \emph{captioning} and \emph{fill-in-the-masks} steps are performed using GPT-4o~\cite{openai2024gpt4}, with the temperature fixed at 0.0 to suppress hallucinations. In the \emph{masking} step, 50\% of nouns and adjectives are randomly replaced with \texttt{[MASK]} tokens. Finally, we synthesize 512 images per style with SDXL-Turbo~\cite{lin2024sdxl}.

\noindent\textbf{Classification Models.} We fine-tune a single linear layer on top of the frozen penultimate layer of CLIP-ViT-B/16~\cite{radford2021learning} using the AdamW~\cite{loshchilov2018decoupled} optimizer with a learning rate of $1\times10^{-4}$ and weight decay of $1\times10^{-2}$. Batch sizes are set to $\min(32, n_r)$ for real images and 512 for synthetic images, respectively. 
\begin{table}[t]
    \centering
    \begin{tabular}{cccccc}
    \toprule
    \multirow{2}{*}{Method} & 
    \multirow{2}{*}{$\bar{D}_{\mathrm{SSIM}}$} & 
    \multirow{2}{*}{$\bar{D}_{\mathrm{LPIPS}}$} & 
    \multicolumn{3}{c}{CMMD$(\downarrow)$}\\
    \cmidrule(lr){4-6}
    & $(\uparrow)$ & $(\uparrow)$ & 1 & 4 & 16 \\
    \midrule
    Class & $\bm{0.424}$ & 0.436 & \multicolumn{3}{c}{\cellcolor{gray!20}3.516} \\
    Caption & 0.398 & 0.418 & 3.203 & 2.146 & $\bm{1.937}$ \\
    Ours & 0.414 & $\bm{0.439}$ & $\bm{2.596}$ & $\bm{2.060}$ & 1.963 \\
    \bottomrule
    \end{tabular}
    \caption{\textbf{Quantitative evaluation of synthetic images.} We assess diversity using SSIM and LPIPS, and quality using CMMD. As Class does not use reference images, its CMMD is constant; the corresponding cells are \colorbox{gray!20}{highlighted}.}
    \label{tab:image_quality_metrics}
\end{table}
\subsection{Few-shot Classification Results}
Tab.~\ref{tab:main_result} presents classification accuracy for each shot. 

\noindent\textbf{Vs. standard augmentations.} MLP without non-generative augmentation outperforms the real-only baselines with standard augmentation in all shot settings except for $n_{\mathrm{shot}}=16$. This result highlights the advantage of MLP, especially under extremely limited supervision.

\noindent\textbf{Vs. Class \& Caption.} Within the column labeled ``N/A'', MLP attains the highest accuracy across all shots, underscoring the importance of balancing fidelity and diversity in generative data augmentation. Class often underperforms relative to the Real Only baseline, indicating the importance of fidelity of synthetic images.

\noindent\textbf{Combination with standard augmentations.} Combining MLP with RandAug. or AugMix further improves performance, suggesting that MLP offers complementary diversity.
\subsection{Analysis on Synthetic Images}
\noindent\textbf{Quantitative Evaluation.} 
We design \emph{reference-conditioned diversity}, $D_{\mathrm{SSIM}}^{(r)}$ and $D_{\mathrm{LPIPS}}^{(r)}$, to assess the diversity among images generated from the same reference image.
\begin{align} 
D_{\text{SSIM}}^{(r)} &= 1-\frac{2}{N(N-1)} \sum_{1 \leq i < j \leq N} 
   \text{SSIM}(x_{i}^{(r)}, x_{j}^{(r)}), \label{eq:dssim} \\
D_{\text{LPIPS}}^{(r)} &= \frac{2}{N(N-1)} \sum_{1 \leq i < j \leq N} 
   \text{LPIPS}(x_{i}^{(r)}, x_{j}^{(r)}), \label{eq:dlpips}
\end{align}
where $N$ denotes the number of generated samples per reference, and $x_{i}^{(r)}$ is the $i$-th image synthesized from the $r$-th reference. For each reference, we compute the Structural Similarity Index Measure (SSIM)~\cite{1284395} and the Learned Perceptual Image Patch Similarity (LPIPS)~\cite{zhang2018perceptual} over all intra-set pairs of generated samples, which quantify pixel-level structural diversity and feature-level semantic diversity, respectively. The final diversity scores, denoted as $\bar{D}_{\text{SSIM}}$ and $\bar{D}_{\text{LPIPS}}$, are obtained by averaging over all $R$ reference images. In our experiments, we prepare $R=624$ reference images evenly sampled from 13 fashion styles from FashionStyle14 dataset. For each reference image, we generate $N=32$ synthesized images and conduct the evaluation. To evaluate the fidelity of synthetic images, we employ the CLIP Maximum Mean Discrepancy (CMMD)~\cite{10656361}, which measures distributional divergence between real and generated images in the CLIP embedding space. For each fashion style, CMMD is computed between the real test samples from the FashionStyle14 dataset and synthetic samples generated under $n_{\mathrm{shot}} = {1, 4, 16}$. The final CMMD score is obtained by averaging across all styles. Tab.~\ref{tab:image_quality_metrics} reports the evaluation results of diversity and fidelity of synthetic images. From the evaluation results of $\bar{D}_{\text{SSIM}}$ and $\bar{D}_{\text{LPIPS}}$, we observe that MLP achieves higher diversity than Caption and comparable diversity to Class. In terms of fidelity, MLP produces distributions closer to real images than those generated by Class, as measured by CMMD. These results highlight the effectiveness of our framework in producing diverse yet style-consistent images, and demonstrate its positive impact on classification performance.

\noindent\textbf{Qualitative Evaluation.} 
Fig.~\ref{fig:wordcloud} shows word frequencies of completions as a word cloud. For the \emph{fairy} style, terms such as ``\emph{cute}'' and ``\emph{pastel}'' appear frequently, reflecting its characteristic aesthetics of cuteness, delicacy, and soft color tones. For the \emph{ethnic} style, words like ``\emph{bohemian},'' ``\emph{floral},'' and ``\emph{brimmed}'' are common, capturing traditional and ethnic designs. These results suggest that GPT tailors completions to the target style. However, as shown in columns 4–5 of Fig.~\ref{fig:mlp_visualization_1}, incorrect completions (\eg, replacing ``\emph{pastel-themed}'' with ``\emph{streetwear}'') can yield images misaligned with the intended style. We leave developing methods to estimate each token’s contribution to style and avoid masking highly influential words as future work.

%% file: sec/4_conclusion.tex
\section{Conclusion}
\label{sec:conclusion}
We propose Masked Language Prompting (MLP), a generative augmentation method for few-shot fashion style recognition. By masking words in reference captions and leveraging LLMs for contextual completion, MLP generates prompts that produce images with high diversity and style consistency. Experiments show that MLP outperforms class-name and caption baselines, and further improves performance when combined with traditional methods. Moreover, analysis on synthetic images shows that MLP enhances diversity over caption-based prompts and better aligns with real data than class-name prompts. In future work, we will construct a multi-label fashion style dataset for more realistic evaluation.

%% file: main.bib
@InProceedings{TakagiICCVW2017,
  author    = {Moeko Takagi and Edgar Simo-Serra and Satoshi Iizuka and Hiroshi Ishikawa},
  title     = {{What Makes a Style: Experimental Analysis of Fashion Prediction}},
  booktitle = ICCVW,
  year      = 2017,
}

@inproceedings{Kiapour2014HipsterWD,
  title={Hipster Wars: Discovering Elements of Fashion Styles},
  author={Mohammad Hadi Kiapour and Kota Yamaguchi and Alexander C. Berg and Tamara L. Berg},
  booktitle=ECCV,
  year={2014},
  volume={8689},
  pages={472--488}
}

@inproceedings{rombach2022high,
  title={High-resolution image synthesis with latent diffusion models},
  author={Rombach, Robin and Blattmann, Andreas and Lorenz, Dominik and Esser, Patrick and Ommer, Bj{\"o}rn},
  booktitle=CVPR,
  pages={10684--10695},
  year={2022}
}

@inproceedings{zhou2023synthetic,
  title={Using Synthetic Data for Data Augmentation to Improve Classification Accuracy},
  author={Yongchao, Zhou and Hshmat, Sahak and Jimmy, Ba},
  booktitle=ICMLW,
  year={2023}
}

@inproceedings{he2023is,
  title={Is synthetic data from generative models ready for image recognition?},
  author={He, Ruifei and Sun, Shuyang and Yu, Xin and Xue, Chuhui and Zhang, Wenqing and Torr, Philip and Bai, Song and Qi, Xiaojuan},
  booktitle=ICLR,
  year={2022}
}

@article{lin2024sdxl,
  title={Sdxl-lightning: Progressive adversarial diffusion distillation},
  author={Lin, Shanchuan and Wang, Anran and Yang, Xiao},
  journal={arXiv preprint arXiv:2402.13929},
  year={2024}
}

@INPROCEEDINGS{simo2016fashion,
  author={Simo-Serra, Edgar and Ishikawa, Hiroshi},
  booktitle=CVPR, 
  title={Fashion Style in 128 Floats: Joint Ranking and Classification Using Weak Data for Feature Extraction}, 
  year={2016},
  volume={},
  number={},
  pages={298--307},
  doi={10.1109/CVPR.2016.39}}

@article{an2023conceptual,
  title={Conceptual framework of hybrid style in fashion image datasets for machine learning},
  author={An, Hyosun and Lee, Kyo Young and Choi, Yerim and Park, Minjung},
  journal={Fashion and Textiles},
  volume={10},
  number={1},
  pages={18},
  year={2023},
}

@article{Ramesh2022HierarchicalTI,
  title={Hierarchical Text-Conditional Image Generation with CLIP Latents},
  author={Aditya Ramesh and Prafulla Dhariwal and Alex Nichol and Casey Chu and Mark Chen},
  journal={arXiv},
  year={2022},
  volume={abs/2204.06125},
  url={https://api.semanticscholar.org/CorpusID:248097655}
}

@inproceedings{yun2019cutmix,
  title={Cutmix: Regularization strategy to train strong classifiers with localizable features},
  author={Yun, Sangdoo and Han, Dongyoon and Oh, Seong Joon and Chun, Sanghyuk and Choe, Junsuk and Yoo, Youngjoon},
  booktitle=ICCV,
  pages={6023--6032},
  year={2019}
}

@inproceedings{
zhang2018mixup,
title={mixup: Beyond Empirical Risk Minimization},
author={Hongyi Zhang and Moustapha Cisse and Yann N. Dauphin and David Lopez-Paz},
booktitle=ICLR,
year={2018},
url={https://openreview.net/forum?id=r1Ddp1-Rb},
}

@inproceedings{
hendrycks2020augmix,
title={AugMix: A Simple Method to Improve Robustness and Uncertainty under Data Shift},
author={Dan Hendrycks* and Norman Mu* and Ekin Dogus Cubuk and Barret Zoph and Justin Gilmer and Balaji Lakshminarayanan},
booktitle=ICLR,
year={2020},
url={https://openreview.net/forum?id=S1gmrxHFvB}
}

@inproceedings{cubuk2020randaugment,
  title={Randaugment: Practical automated data augmentation with a reduced search space},
  author={Cubuk, Ekin D and Zoph, Barret and Shlens, Jonathon and Le, Quoc V},
  booktitle=CVPRW,
  pages={702--703},
  year={2020}
}

@inproceedings{cubuk2019autoaugment,
  title={Autoaugment: Learning augmentation strategies from data},
  author={Cubuk, Ekin D and Zoph, Barret and Mane, Dandelion and Vasudevan, Vijay and Le, Quoc V},
  booktitle=CVPR,
  pages={113--123},
  year={2019}
}

@inproceedings{radford2021learning,
  title={Learning transferable visual models from natural language supervision},
  author={Radford, Alec and Kim, Jong Wook and Hallacy, Chris and Ramesh, Aditya and Goh, Gabriel and Agarwal, Sandhini and Sastry, Girish and Askell, Amanda and Mishkin, Pamela and Clark, Jack and others},
  booktitle=ICML,
  pages={8748--8763},
  year={2021}
}

@inproceedings{
dunlap2023diversify,
title={Diversify Your Vision Datasets with Automatic Diffusion-based Augmentation},
author={Lisa Dunlap and Alyssa Umino and Han Zhang and Jiezhi Yang and Joseph E. Gonzalez and Trevor Darrell},
booktitle=NIPS,
year={2023},
url={https://openreview.net/forum?id=9wrYfqdrwk}
}

@inproceedings{kim2024datadream,
  title={Datadream: Few-shot guided dataset generation},
  author={Kim, Jae Myung and Bader, Jessica and Alaniz, Stephan and Schmid, Cordelia and Akata, Zeynep},
  booktitle=ECCV,
  pages={252--268},
  year={2024},
}

@inproceedings{wang2024attributed,
    title={Attributed Synthetic Data Generation for Zero-shot Image Classification},
    author={Shijian Wang and Linxin Song and Ryotaro Shimizu and Masayuki Goto and Hanqian Wu},
    booktitle={Synthetic Data for Computer Vision Workshop @ CVPR 2024},
    year={2024},
    url={https://openreview.net/forum?id=k4Xnh0EPus}
}

@article{openai2024gpt4,
  author = {OpenAI},
  title = {GPT-4 Technical Report},
  journal={arXiv preprint arXiv:2303.08774},
  year = 2023
}

@inproceedings{
loshchilov2018decoupled,
title={Decoupled Weight Decay Regularization},
author={Ilya Loshchilov and Frank Hutter},
booktitle=ICLR,
year={2019},
url={https://openreview.net/forum?id=Bkg6RiCqY7},
}

@inproceedings{
zhang2023expanding,
title={Expanding Small-Scale Datasets with Guided Imagination},
author={Yifan Zhang and Daquan Zhou and Bryan Hooi and Kai Wang and Jiashi Feng},
booktitle=NIPS,
year={2023},
url={https://openreview.net/forum?id=82HeVCqsfh}
}

@INPROCEEDINGS{10656361,
  author={Jayasumana, Sadeep and Ramalingam, Srikumar and Veit, Andreas and Glasner, Daniel and Chakrabarti, Ayan and Kumar, Sanjiv},
  booktitle=CVPR, 
  title={Rethinking FID: Towards a Better Evaluation Metric for Image Generation}, 
  year={2024},
  volume={},
  number={},
  pages={9307--9315},
  doi={10.1109/CVPR52733.2024.00889}}

@ARTICLE{1284395,
  author={Zhou Wang and Bovik, A.C. and Sheikh, H.R. and Simoncelli, E.P.},
  journal=TIP, 
  title={Image quality assessment: from error visibility to structural similarity}, 
  year={2004},
  volume={13},
  number={4},
  pages={600--612},
  doi={10.1109/TIP.2003.819861}}

@inproceedings{zhang2018perceptual,
  title={The Unreasonable Effectiveness of Deep Features as a Perceptual Metric},
  author={Zhang, Richard and Isola, Phillip and Efros, Alexei A and Shechtman, Eli and Wang, Oliver},
  booktitle=CVPR,
  year={2018}
}

@inproceedings{XuewenECCV20Fashion,
Author = {Xuewen Yang and Heming Zhang and Di Jin and Yingru Liu and Chi-Hao Wu and Jianchao Tan and Dongliang Xie and Jue Wang and Xin Wang},
Title = {Fashion Captioning: Towards Generating Accurate Descriptions with Semantic Rewards},
booktitle = ECCV,
Year = {2020}
}
